\documentclass[11pt]{article}

\usepackage{ACL2023}

\usepackage{times}
\usepackage{latexsym}

\usepackage[T1]{fontenc}
\usepackage[utf8]{inputenc}

\usepackage{microtype}
\usepackage{multirow}
\usepackage{graphicx}
\usepackage{enumitem}
\usepackage{fancyhdr}
\usepackage{inconsolata}
\usepackage{pgfplots}
\pgfplotsset{compat=1.17}
\newcommand\blfootnote[1]{%
  \begingroup
  \renewcommand\thefootnote{}\footnote{#1}%
  \addtocounter{footnote}{-1}%
  \endgroup
}

\title{SentMix-3L: A Bangla-English-Hindi Code-Mixed Dataset for \\ Sentiment Analysis}

\author{Md Nishat Raihan\textsuperscript{*}, Dhiman Goswami\textsuperscript{*}, Antara Mahmud \\  \textbf{Antonios Anastasopoulos}, \textbf{Marcos Zampieri} \\
        George Mason University, USA \\
        \texttt{\{mraihan2, dgoswam, amahmud4, antonis, mzampier\}@gmu.edu} \\
        }

\begin{document}
\maketitle
\begin{abstract}

Code-mixing is a well-studied linguistic phenomenon when two or more languages are mixed in text or speech. Several datasets have been build with the goal of training computational models for code-mixing. Although it is very common to observe code-mixing with multiple languages, most datasets available contain code-mixed between only two languages. In this paper, we introduce SentMix-3L, a novel dataset for sentiment analysis containing code-mixed data between three languages Bangla, English, and Hindi. We carry out a comprehensive evaluation using SentMix-3L. We show that zero-shot prompting with GPT-3.5 outperforms all transformer-based models on SentMix-3L. \blfootnote{*These two authors contributed equally to this work.}
\end{abstract}

\section{Introduction}

Code-mixing and code-switching are very commonly observed in both text and speech. Code-mixing means the practice of using words from multiple languages within a single utterance, sentence, or discourse, and code-switching refers to the deliberate alteration between multiple languages within the same context \cite{thara2018code}. The first case is spontaneous and the second case is purposeful. However, both are widely observed in bilingual and multilingual communities. 

According to \citet{anastassiou2017factors}, several factors are behind these two phenomena, which include social, convenience, linguistic, and cognitive reasons. Socially, this often serves as a sign of group identity which allows individuals to navigate multiple social and cultural affiliations. In terms of linguistics, it is a very common scenario to not be able to find any word for a specific term in one language, whereas another word from another language can help to communicate better. Additionally, there are several cases even in a monolingual community, when Code-mixing might be the convenient way to express something. 

In most occurrences, code-mixing is bilingual. In an early research, \citet{byers2013bilingualism} states that, it is very likely that by the year 2035, over half of the children enrolled in kindergarten will have grown up speaking a language other than English. Another study conducted by \citet{jeffery2020language} shows that it is very common in European countries like Germany, Spain, and Italy to use bilingualism in practice. However, in cosmopolitan cities and areas like New York, London, Singapore, and others, code-mixing with three or even more languages is fairly common. This is also observed in countries like Luxembourg, and regions such as West Bengal, and South-East India where more than two languages are commonly used on a daily basis.

Several research works have been conducted on building code-mixed datasets and performing several downstream tasks on such datasets. These datasets include both synthetic and natural ones. However, most of them are bilingual in nature. In this paper, we present SentMix-3L, a Bangla-English-Hindi dataset annotated for sentiment analysis. 

The main contributions of our work are as follows:

\begin{itemize}
    \item We introduce SentMix-3L, a novel three-language code-mixed test dataset with gold standard labels in Bangla-Hindi-English for the task of Sentiment Analysis, containing 1,007 instances.\footnote{https://github.com/LanguageTechnologyLab/SentMix-3L}
    \item We provide a comprehensive experimental analysis with several monolingual, bilingual, and multilingual models on SentMix-3L.
\end{itemize}

\noindent We are presenting this dataset exclusively as a test set due to the unique and specialized nature of the task. Such data is very difficult to gather and requires significant expertise to access. The size of the dataset, while limiting for training purposes, offers a high-quality testing environment with gold-standard labels that can serve as a benchmark in this domain. Given the scarcity of similar datasets and the challenges associated with data collection, SentMix-3L provides an important resource for the rigorous evaluation of text-based models, filling a critical gap in multi-level Code-mixing research. In our experiments, we also prepare a synthetic train and a development dataset to evaluate several models.

\section{Related Work}
% \begin{itemize}
%     \item Bn-En/ Bn-Hi relevant code mixed dataset or works.
%     \item Code Mixed (any) Sentiment Task
% \end{itemize}

There have been some works conducted on Bangla-English, Hindi-English, and Bangla-Hindi Code-mixing and Code-switching separately. Most of them are case studies and surveys that show the common occurrences of Bangla-English \cite{alam2006code, hasan2015reviewing, hossain2015case, begum2013code, mahbub2016sociolinguistic}, Hindi-English \cite{singh1985grammatical, bali2014borrowing, thara2018code} and Bangla-Hindi \cite{ali2019effects, jose2020survey} Code-mixing in a wide variety of areas and situations.

Few works are done on sentiment analysis tasks for these types of cases. The work of \citet{khan2022sentiment} presents a Bangla-English Sentiment Analysis dataset primarily related to COVID-19. Their dataset is called CoVaxBD and it contains 1113 samples. Their experiments show that the best result of the dataset is generated by BERT with a development accuracy of 97.3\%. However, a lot of the data are purely in Bangla and they only experiment using the BERT and multilingual BERT model \cite{devlin2019bert} while only providing development accuracy as the performance metric. Another recent work by \citet{tareq2023data}, consists of 18,074 Bangla-English Code-mixed data from online for the purpose of sentiment analysis. They augment the dataset using their own approach and achieve their best result of an 87\% weighted F1 score by implementing XGBoost with Fasttext embedding. Their experiments lack the evaluation of transformer-based models.

An early work by \citet{sitaram2015sentiment} focuses on Hindi-English Code-mixing dataset for Sentiment Analysis. However, it contains 345 data samples in total and only 180 of them are code-mixed. They get their best results of 91.01\% accuracy using Recursive Neural Tensor Network (RNTN). \citet{joshi2016towards} compile a dataset of 3879 texts and get their best results using a subword-LSTM approach. Also, the work of \cite{yadav2020bi} includes a dataset of 6357 texts and Bi-LSTM helps them to get their optimal result. However, none of these works present how the transformer-based models perform on their datasets.

%However, no work has been done specifically on Bangla-Hindi Code-mixing. However, there are a few works on the same task for several Indian languages altogether including Bangla and Hindi \cite{patra2018sentiment, ahmad2022machine}.

In summary, there are no works or datasets on sentiment analysis for code-mixed Bangla-English-Hindi altogether. %Hence, we are also unable to determine how the State-of-the-art models would perform on such datasets. 
SentMix-3L is a novel addition in this particular domain of research.

\section{The \textit{SentMix-3L} Dataset}
% \begin{itemize}
%     \item Data Collection
%     \item Dataset Annotation
%     \item Dataset Statistics
%     \item Dataset Card
%     \item Synthetic Dataset
% \end{itemize}
In generating the dataset, we choose a controlled data collection method, asking the volunteers to freely contribute data in Bangla, English, and Hindi. This decision stems from several challenges of extracting such specific code-mixed data from the vast corpus available on social media or other online platforms. While the data are not rare, identifying and isolating them from large, unstructured corpora is a very labor-intensive and error-prone process. Our approach ensures data quality and sidesteps the ethical concerns associated with using publicly available online data. Such types of datasets are often used when it is very difficult to mine them from existing corpora. As examples, for fine-tuning LLMs on instructions and conversations, semi-natural datasets like \citet{dolly2023} and \citet{awesome_instruction_datasets} have become popular.

% \begin{table*} [!h]
% \centering
% \begin{tabular}{lccccc}
% \hline
% \textbf{No. of Tokens} & \textbf{All} & \textbf{Bangla} & \textbf{English} & \textbf{Hindi} & \textbf{Other}\\
% \hline
% Total & 89494 & 32133 & 5998 & 15131 & 36232\\
% Unique & 19686 & 8167 & 1073 & 1474 & 9092\\
% Maximum & 188 & 72 & 18 & 36 & 80\\
% Minimum & 36 & 10 & 4 & 2 & 11\\
% Average & 88.87 & 31.91 & 5.96 & 15.03 & 35.98\\
% Std. Dev. & 19.19 & 8.39 & 2.94 & 5.81 & 9.70\\
% \hline
% \end{tabular}
% \caption{Data Card}
% \label{tab:data_card}
% \end{table*}

\paragraph{Data Collection} %To gather a text dataset that contains Bangla-English-Hindi Code-mixing for all the samples from any social media or online corpus is very challenging. There is no standard way to mine such data in an efficient way. Hence, we gather 
A group of 10 undergraduate students who are fluent in all 3 languages in all four language skills - listening, reading, writing, and speaking. We ask each of them to prepare 250 to 300 social media posts or tweets. They are allowed to use any language including Bangla, English, and Hindi to prepare posts on several daily topics like politics, sports, education, social media rumors, etc. We also ask them to switch languages if and wherever they feel comfortable doing it. The inclusion of emojis, hashtags, and transliteration are also encouraged. The students had the flexibility to prepare the data as naturally as possible. Upon completion of this stage, we filter 1863 samples that contain at least one word or subword from each of the three languages using langdetect \cite{langdetect} an open-sourced Python tool for language detection. 

\paragraph{Data Annotation} We annotate the dataset in two steps to prepare high-quality labels for the dataset. First, we recruit three students from social science, computer science, and linguistics as annotators who are also fluent in all 3 languages in all four language skills. They annotate all the 1863 samples with one of the three labels (Positive, Neutral, and Negative) with a raw agreement of 65.3\%. We only take these 1182 data, where all three annotators agree on the labels. Second, we gather a second group of annotators consisting of two NLP researchers with the same level of fluency and skills. After their annotation, we calculate a raw agreement of 0.85, a Cohen Kappa score of 0.78 and only keep the data where both annotators agree. After the two stages, we end up with a total of 1007 data. 

\begin{table} [!h]
\centering
\begin{tabular}{lcc}
\hline
\textbf{Label} & \textbf{No. of Data} & \textbf{Percentage}\\
\hline
Positive & 420 & 41.71\% \\
Neutral & 234 & 23.24\% \\
Negative & 353 & 35.05\% \\
\hline
Total & 1,007 & 100\% \\
\hline
\end{tabular}
\caption{Label distribution in SentMix-3L.}
\label{tab:label_distribution}
\end{table}

\paragraph{Dataset Statistics} A detailed description of the dataset statistics is provided in Table \ref{tab:data_card}. Since the dataset was generated by people whose first language is Bangla, we observe that the majority portion of the tokens in the dataset is in Bangla. It is also observed that all the data instances contain tokens from all three languages. There are several \textit{Other} tokens in the dataset that are not from Bangla, English, or Hindi language. These tokens contain transliterated words and misspelled words as well as emojis and hashtags. We have Positive, Neutral, and Negative labels for the 1007 data. The label distribution is shown in table \ref{tab:label_distribution}. 

\begin{table} [!h]
\centering
\scalebox{.82}{
\begin{tabular}{l|c|cccc}
\hline
 & \textbf{All} & \textbf{Bangla} & \textbf{English} & \textbf{Hindi} & \textbf{Other}\\
\hline
Tokens & 89494 & 32133 & 5998 & 15131 & 36232\\
Types & 19686 & 8167 & 1073 & 1474 & 9092\\
% Max. in instance & 173 & 62 & 20 & 47  & 93\\
% Min. in instance & 41 & 4 & 3 & 2 & 8\\
Avg & 88.87 & 31.91 & 5.96 & 15.03 & 35.98\\
Std Dev & 19.19 & 8.39 & 2.94 & 5.81 & 9.70\\
\hline
\end{tabular}
}
\caption{SentMix-3L Data Card. The row {\em Avg} represents the average number of tokens with its standard deviation in row {\em Std Dev}.}
\label{tab:data_card}
\end{table}

\paragraph{Synthetic Train and Development Set} We present SentMix-3L as a test dataset, hence for experimental purposes, we build a synthetic train and development set that contains Code-mixing for Bangla, English, and Hindi. We originally take the Amazon Review Dataset \cite{ni2019justifying} as seed data and pick 100K data instances randomly. The dataset labels are ratings on a 1 to 5 scale. We convert them into Positive (rating > 3), Neutral (rating = 3), and Negative (rating < 3) for our task. We carefully choose an equal number of instances for Positive, Neutral, and Negative labels. We then use two separate methodologies called \textit{Random Code-mixing Algorithm} by \citet{krishnan2021multilingual} and \textit{r-CM} by \citet{santy2021bertologicomix} to generate the synthetic Code-mixed dataset. 

\section{Experiments}

%For the experimental setup, this tri-lingual code-mixed sentiment analysis task is divided into four major parts. We use monolingual, bilingual, multilingual transformer-based models and prompting for the detailed analysis of the data. 
%For the experiments, A100 GPU and 80 GB system memory are used for all experiments, and the best hyper-parameters are chosen for all the models by performing an empirical analysis. We use the synthetic data as the train set and the natural data as test set.

\paragraph{Monolingual Models} We use 5 monolingual models  DistilBERT \cite{DBLP:journals/corr/abs-1910-01108}, BERT \cite{devlin2019bert}, BanglaBERT \cite{kowsher2022bangla}, roBERTa \cite{liu2019roberta}, HindiBERT \cite{nick_doiron_2023} for this experiment. Here BanglaBERT is specifically trained on only Bangla and HindiBERT on only Hindi language. The rest of the models are trained in the English language only.

\paragraph{Bilingual Models} BanglishBERT \cite{bhattacharjee-etal-2022-banglabert} and HingBERT \cite{nayak-joshi-2022-l3cube} is used as bilingual models which are trained on both Bangla-English and Hindi-English respectively thus effective for the purpose of code mixing tasks including where any two of these languages are involved.

\paragraph{Multilingual Models} We use mBERT \cite{devlin2019bert} and XLM-R \cite{conneau2019unsupervised} as multilingual models which are respectively trained on 104 and 100 languages. These two models are very effective while we are working in a trilingual  Bangla, Hindi and English domain. Moreover, we use IndicBERT \cite{kakwani2020indicnlpsuite} and MuRIL \cite{khanuja2021muril} which covers 12 and 17 Indian languages respectively including Bangla-English-Hindi. Thus these two models and the respective list of languages justify the inclusion of them for our targeted tri-lingual code-mixing task. We also perform hyper-parameter tuning while using all the models to prevent overfitting and ensure optimal F1 score.

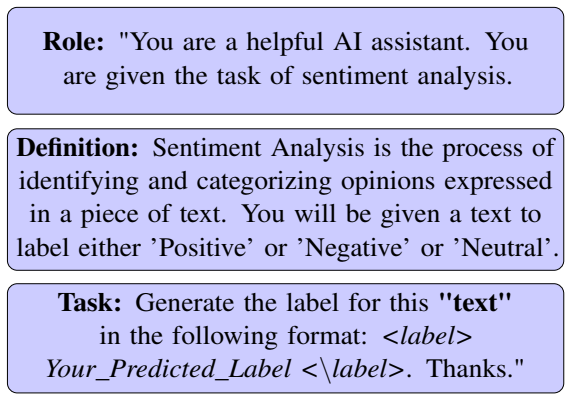
\begin{figure}[h]
\centering
\scalebox{.92}{
\begin{tikzpicture}[node distance=1cm]
    % Styles for nodes
    \tikzstyle{block} = [rectangle, draw, fill=blue!20, text width=\linewidth, text centered, rounded corners, minimum height=4em]
    \tikzstyle{operation} = [text centered, minimum height=1em]
    % Nodes
    \node [block] (rect1) {\textbf{Role:}{ "You are a helpful AI assistant. You are given the task of sentiment analysis. }};
    \node [operation, below of=rect1] (plus1) {};
    \node [block, below of=plus1] (rect2) {\textbf{Definition:}{ Sentiment Analysis is the process of identifying and categorizing opinions expressed in a piece of text. You will be given a text to label either 'Positive' or 'Negative' or 'Neutral'. }};
    \node [operation, below of=rect2] (plus2) {};
    \node [block, below of=plus2] (rect3) {\textbf{Task:}{ Generate the label for this \textbf{"text"} in the following format: \textit{<label> Your\_Predicted\_Label <$\backslash$label>}. Thanks."}};
\end{tikzpicture}
}
\caption{Sample GPT-3.5 prompt.}
\label{fig:prompt1}
\end{figure}

\paragraph{Prompting}  We use prompting with GPT-3.5-turbo model \cite{openai2023gpt35turbo} from OpenAI for this task. We use the API for zero-shot prompting (see Figure \ref{fig:prompt1}) and ask the model to label the test set. \\

Additionally, we run the same experiments separately on synthetic and natural datasets splitting both in a 60-20-20 way for training, evaluating, and testing purposes.

\section{Results}

In this experiment, synthetic data is used as train set, and natural data is used as test set. The F1 scores of monolingual models range from 0.47 to 0.55 where BERT performs the best. Among the two bilingual models BanglishBERT scores 0.56 which is better than HingBERT. XLM-R is the best multilingual model with an F1 score of 0.59. On the other hand, a zero shot prompting technique on GPT 3.5 turbo performs the best with a 0.62 weighted F1 score. These results are available in Table \ref{Results1}. %and Appendix \ref{Appendix_A}.

\begin{table} [!h]
\centering
\scalebox{.95}{
\begin{tabular}{lcc}
\hline
\textbf{Models}  & \textbf{Weighted F1 Score} \\
\hline
GPT 3.5 Turbo & \textbf{0.62} \\
XLM-R & 0.59 \\
BanglishBERT & 0.56 \\
mBERT & 0.56 \\
BERT & 0.55 \\
roBERTa & 0.54 \\
MuRIL & 0.54 \\
IndicBERT & 0.53 \\
DistilBERT & 0.53 \\
HindiBERT & 0.48 \\
HingBERT & 0.47 \\
BanglaBERT & 0.47 \\
\hline
\end{tabular}
}
\caption{Weighted F-1 score for different models: training on synthetic, testing on natural data.}
\label{Results1}
\end{table}

\noindent We perform the same procedure by using synthetic data for both the training and testing where MuRIL and XLM-R with 0.77 F1 score are the best-performing models. These results are available in Table \ref{Results2}.

%Lastly, we use natural data for both the training and testing where BanglishBERT scores 0.90 F1 score which is the best among all the models. These results are available in Appendix \ref{Appendix_B}.

\begin{table} [!h]
\centering
\scalebox{.95}{
\begin{tabular}{lcc}
\hline
\textbf{Models}  & \textbf{Weighted F1 Score} \\
\hline
MuRIL & \textbf{0.77} \\
XLM-R & \textbf{0.77} \\
mBERT & 0.74 \\
BanglishBERT & 0.72 \\
BERT & 0.68 \\
HingBERT & 0.67 \\
DistilBERT & 0.66 \\
roBERTa & 0.66 \\
IndicBERT & 0.66 \\
BanglaBERT & 0.60 \\
HindiBERT & 0.59 \\
\hline
\end{tabular}
}
\caption{Weighted F-1 score for different models: training and testing on synthetic data.}
\label{Results2}
\end{table}

\subsection{Error Analysis}

We observe \textit{Other} tokens in almost 40\% of the whole dataset, as shown in Table \ref{tab:data_card}. These tokens occur due to transliteration which poses a challenge for most of the models since not all of the models are pre-trained on transliterated tokens. BanglishBERT did better than HingBERT since it recognizes both Bangla and English tokens and the total number of tokens for Hindi-English is less than Bangla-English tokens (see Table \ref{Results1}). Also, misspelled words and typos are also observed in the datasets, making the task even more difficult. Some examples are available in Appendix \ref{sec:appendix_a} which are classified wrongly by all the models.

\section{Conclusion and Future Work}

In this paper, we presented SentMix-3L, a Bangla-English-Hindi code-mixed offensive language identification dataset containing 1,007 instances. We also created 100,000 synthetic data in the same three languages for training. We evaluated various monolingual models on these two datasets. Our results show that prompting GPT3.5 generates the best result on SentMix-3L. When using synthetic data for both training and testing, multilingual models such as mBERT and XLM-R perform well. 
In the future, we would like to expand SentMix-3L so that it can serve as both training and testing data. Additionally, we are working on pre-training Bangla-English-Hindi trilingual code-mixing models for offensive language identification. 

\section*{Acknowledgements}
We thank the annotators who helped us with the annotation of SentMix-3L. We further thank the anonymous workshop reviewers for their insightful feedback. 
Antonios Anastasopoulos is generously supported by NSF award IIS-2125466.

\section*{Limitations}

Although most datasets for the downstream tasks are scraped from social media posts in the real world, in our case these data instances are generated in a semi-natural manner, meaning that they were generated by people but not scraped from social media directly. This was done due to the complexity of extracting contents that contain a specific 3 language code-mixing in them. Also, the dataset is comparatively smaller in size, since it is costly to generate data by a specific set of people who are fluent in all 3 target languages.

\section*{Ethics Statement}

The dataset introduced in this paper, which centers on the analysis of offensive language in Bangla-English-Hindi code-mixed text, adheres to the  \href{https://www.aclweb.org/portal/content/acl-code-ethics}{ACL Ethics Policy} and seeks to make a valuable contribution to the realm of online safety. SentMix-3L serves as an important resource for opinion analysis of online content. Moreover, the contributors and annotators of the dataset are paid respectable remuneration for their efforts towards this dataset.

% Entries for the entire Anthology, followed by custom entries
\bibliography{custom}
\bibliographystyle{acl_natbib}

\appendix
\section{Examples of Misclassified Instances}
\label{sec:appendix_a}
\includegraphics[width = \linewidth]{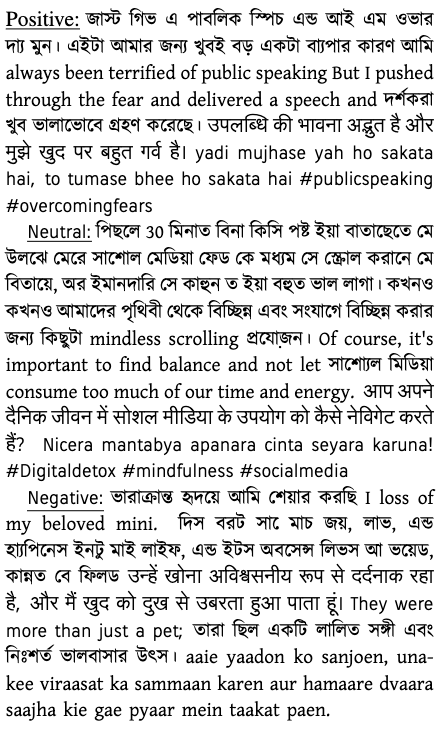}

\end{document}